\documentclass[sigconf]{acmart}
\AtBeginDocument{%
  \providecommand\BibTeX{{%
    \normalfont B\kern-0.5em{\scshape i\kern-0.25em b}\kern-0.8em\TeX}}}

\usepackage{comment}

\begin{document}

\title{A domain adaptive deep learning solution for scanpath prediction of paintings}

\author{Mohamed Amine Kerkouri* }       
\email{mohamed-amine.kerkouri@univ-orleans.fr}
\thanks{ * Equal contribution; the order of first authors was randomly selected}
\affiliation{%
  \institution{Université d'Orléans}
  \streetaddress{12 Rue de Blois}
  \city{Orléans}
  \country{France}
  \postcode{45100}
}

\author{Marouane Tliba*}
\email{marouane.tliba@univ-orleans.fr}
\affiliation{%
  \institution{Université d'Orléans}
  \streetaddress{12 Rue de Blois}
  \city{Orléans}
  \country{France}
  \postcode{45100}
}

\author{Aladine Chetouani}
\affiliation{%
  \institution{Université d'Orléans}
  \streetaddress{12 Rue de Blois}
  \city{Orléans}
  \country{France}
  \postcode{45100}
}

\author{Alessandro Bruno}
\affiliation{%
  \institution{Bournemouth University}
  \streetaddress{Talbot Campus, Fern Barrow}
  \city{Poole}
  \country{United Kingdom}
  \postcode{BH125BB}
}

\renewcommand{\shortauthors}{KERKOURI et al.}

\begin{abstract}
Cultural heritage understanding and preservation is an important issue for society as it represents a fundamental aspect of its identity. 
Paintings represent a significant part of cultural heritage, and are the subject of study continuously. However, the way viewers perceive paintings is strictly related to the so-called HVS (Human Vision System) behaviour. 
This paper focuses on the eye-movement analysis of viewers during the visual experience of a certain number of paintings. 
In further details, we introduce a new approach to predicting human visual attention, which impacts several cognitive functions for humans, including the fundamental understanding of a scene, and then extend it to painting images.  
The proposed new architecture ingests images and returns scanpaths, a sequence of points featuring a high likelihood of catching viewers' attention.  
We use an FCNN (Fully Convolutional Neural Network), in which we exploit a differentiable channel-wise selection and Soft-Argmax modules. We also incorporate learnable Gaussian distributions onto the network bottleneck to simulate visual attention process bias in natural scene images. Furthermore, to reduce the effect of shifts between different domains (i.e. natural images, painting), we urge the model to learn unsupervised general features from other domains using a gradient reversal classifier. The results obtained by our model outperform existing state-of-the-art ones in terms of accuracy and efficiency.
\end{abstract}




\maketitle

\section{Introduction}

Cultural heritage encompasses the habits, traditions, artifacts, and artistic expressions produced by a community and carried down through generations \cite{HARTMANN2020369}. Thus representing an integral part of their communal identity which should be preserved.   

Paintings represent an important part of the artistic heritage, it is a picture created by putting paint on a surface, or the activity or skill of creating pictures by using paint\footnote{\url{https://dictionary.cambridge.org/dictionary/english/painting}}. They depict a pictorial recounting of events, ideas, and concepts of the environment, era, and tendencies of the artist. Thus, the preservation and understanding of such heritage are important as they are an integral part of human history.   

Understanding this type of media is directly linked to the visual perception mechanisms of the human visual system (HVS) and as human eyes have a resolution of more than 500 MP and receive about 10 Gigabits of information per second \cite{brain}.

Due to the enormous quantity of data to be processed, HVS needs to filter the most relevant areas in a visual scene while discarding the least meaningful ones from a perceptual perspective.  

In greater detail, eye movements allow filtering out non-relevant information by attending to a region of the visual field rather than others. That mechanism unburdens the cognitive load and increases the efficiency of the visual perception system. 
Findings revealed that low-level visual features (color, edges, texture, intensity, contrast, etc) catch eye movements in the early stages of image observation. In contrast, semantic visual features (faces, text, humans, objects, etc.) play a critical role in a later stage of the visual experience \cite{feature_integration}.

The aforementioned mechanism can be categorized into 'covert' and 'overt' attention. The former focuses on optimizing the processing activity in the brain cortex. The latter has a strict relation with eye movements that bring regions of interest of the visual field onto the fovea. Eye-trackers capture eye movements and allow drawing the gaze trajectory of viewers while exploring a specific visual stimulus. The scanpath consists of fixation points and saccades. Saccades are fast eye movements with no visual perception between fixation points \cite{10.1145/1743666.1743717}.

Fixations of multiple observers for a specific stimulus (an image or a video sequence) are generally represented with the so-called fixation point maps. These maps can be smoothed using a Gaussian kernel representing $1^\circ$ of visual angle to give a spatial distribution of eye movements. Furthermore, they provide a probability heat map called a saliency map where each pixel has a normalized probability value of grabbing the viewer's attention. 

The main goal of saliency models is to generate saliency maps as close as possible to fixation point maps \citep{satsal}, \citep{Tliba_2022_CVPR}. 

Attention can also be classified using the "neural processing path" it uses for the mechanism. "Bottom-up Attention" is mainly directed by the external stimuli features, like the one aforementioned previously. This means that the intrinsic properties and features of the images reign over the mechanism, and this correlates with the theory that states that attention is pre-processing operation. This type is unintentional, fast and instinctual.

Meanwhile, the other type is called "Top-down Attention". This type is influenced by other highly cognitive processes originating from the prefrontal cortex and the long-term memory, it relates to the theory stating that attention is a post-processing operation. This mechanism is intentional, slow, and premeditated.

The output of saliency models is used in several topics such as image quality assessment \cite{QAChetouaniICIP2018,Chetouani20EUSIPCO,Ilyass19NCAA,PR20Ilyass,SPIC20Chetouani}, image and video compression \cite{saliencyComp1}, image captioning and description \cite{saliencyCapt1}, image search and retrieval \cite{retrivalSal}, image enhancement for people with CVD (Colour Vision Deficiency) \cite{bruno2019image}, saliency led painting restoration \cite{painting_restor} and so forth \cite{info19Hamidi}.

Scientific research in saliency prediction was pioneered by the seminal work of \cite{KochUllman}, implemented by \cite{Itti} using a multi-scale model reliant on the extraction of low-level features (i.e. color, intensity, and orientation). 

Alongside visual saliency, scanpath prediction has gained the interest of researchers lately. The winner-take-all (WTA) module presented by \cite{Itti} was the first to predict scanpaths. They extracted the points with the highest saliency value as fixation points and inhibited the region surrounding that point. 
In \cite{lemeur}, the authors proposed a stochastic approach to generate scanpaths. They predicted saliency maps and modeled the probabilities of several biases (i.e. saccade amplitudes and saccade orientations). 

The authors of \cite{G-Eymol} analogized the saliency map with a 2D gravitational field affecting the trajectory of a mass representing the gaze.  
The authors in \cite{pathgan} developed a Long-Short-Term-Memory (LSTM) network. They combined it with a conditional GAN (Generative Adversarial Network) architecture to predict the scanpath of a given visual stimulus. 
A framework employing foveal saliency maps, temporal duration, and a model of the Inhibition of Return (IoR) to predict scanpath was proposed in \cite{DCSM}. 

The authors of \cite{SALYPATH} introduced an end-to-end model which simultaneously predicts the scanpath and the saliency map of an image. 
Then, in \cite{simple_scanpath} they used a fully convolutional architecture of known computer vision models in a regressive manner to predict scanpaths.

\begin{figure*}
    \centering
    \includegraphics[width= 0.9\linewidth, height = 70mm]{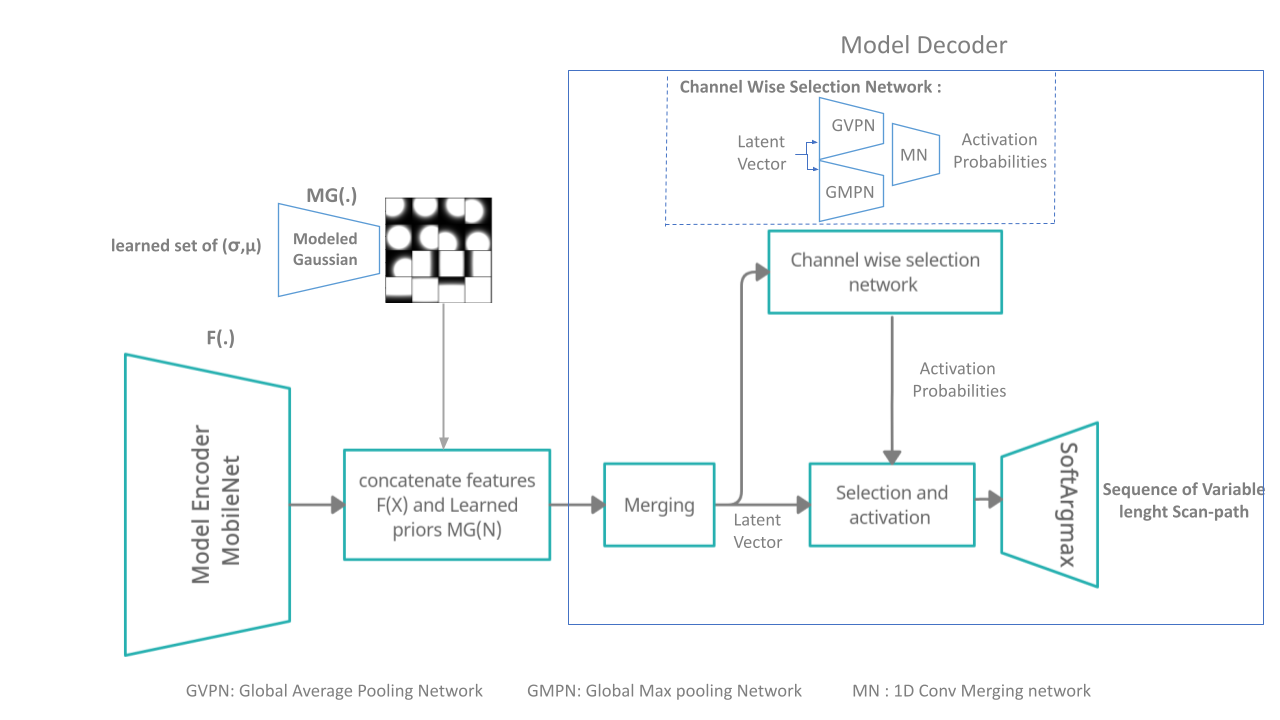}
    \caption{\label{figure:arch}General architecture of the proposed method.}
    \vspace{-3mm}
\end{figure*}

In this paper, we propose a novel efficient method for vector to sequence modeling (Image to scanpath) using fully Convolutional Neural Networks.
Because images originating from different domains ( i.e. natural scenes, art, synthetic, ... etc) exhibit different biases related to the domains themselves and at the same time visual perception tasks also have their own subjective biases. It becomes hard for models to model relevant features specific to the images while ignoring these types of noisy distributions.  
To that end, We investigate two sorts of domain adaptation techniques one related to the downstream task priors biases formulation and the other focusing on feature extraction generalization also known as domain representation alignment, over different data distribution domains and employ this technique to improve the prediction on paintings dataset. 

The main contributions of this work are summarized as it follows: 
\begin{itemize}
  \item Propose a novel and efficient deep model for scanpath prediction.
  \item Investigate domain adaptation on scanpath prediction by learning human visual attention bias and use unsupervised adversarial domain adaption to generalize our model prediction on other image domains ( paintings, etc.).
  \item Conduct an extensive qualitative and quantitative experimental campaign to the proposed method's effectiveness.
 
\end{itemize}

The rest of the paper is organized as follows: in Section \ref{sec:method}, we describe the components of the proposed model architecture as well as the training protocol and the generalization process using the unsupervised domain adaptation. In section \ref{sec:results}, the results and the evaluation of the model are detailed. Section \ref{sec:conclusion} ends the paper with conclusions.

\section{Proposed Method}
\label{sec:method}
This work aims to introduce a new framework for scanpath prediction and demonstrate the effectiveness of unsupervised domain adaptation on the task. 
The general architecture of the proposed model is graphically depicted in Fig.\ref{figure:arch}. We first use MobileNet as the backbone for feature extraction. Secondly, learnable domain prior bias maps are concatenated to the output of the feature extractor and then passed onto a fully convolutional decoder.
The latter is composed of several modules that aim to extract specific characteristics relative to the final prediction. Global pooling features are extracted from the maps resulting from  the merging module and then passed to a fully connected network in order to predict a probability vector. The latter is then discretized into a binary mask related to the relevant predicted fixation points of the scanpath. The probability vector is used as a channel-wise weighting vector to accentuate the importance of specific channels compared to others. Finally, the resulting weighted feature maps are passed through a Soft-ArgMax function, where each fixation point coordinate is predicted from each feature map.\footnote{code will  soon be availble at : \url{https://github.com/Submit-code/ScanpathDomain}}    
    
\subsection{Feature extractor}

Pixel-wise computations can be labor-intensive on dimensional complexity and spatial correlation. CNNs are used to model their features before extracting them. Different architectures were proposed and evolved for this end \cite{Mahmoudi2020ICIP,MAHMOUDI2020PRL}. 
Here, we use MobileNet \cite{mobileNet} to extract features from the input images. It is a lightweight CNN built for use in mobile and embedded applications. It introduced depth-wise separable convolutions and point-wise convolutions to decrease the number of features.  This network helps the architecture to remain lightweight and is easier to train with the amount of data  we have.   
    
    \subsection{Priority maps and biases}
Saliency in images exhibits a phenomenon called center bias, where the gazes of observers are more oriented toward the center of the image compared to the edges and corners of the stimulus. That is mainly due to the tendency for relevant semantic information to be located around central regions of images. 
The center bias phenomenon can be modeled using a probability map representing a 2-dimensional Gaussian function as shown below: 

\begin{equation}
    f(x,y) = \frac{1}{2\pi \sigma_x \sigma_y} \exp^{-( (\frac{x-\mu_x}{2\sigma_x}) + (\frac{y-\mu_y}{2\sigma_y}))}
    \label{eq:gaussian2d} 
\end{equation}
where $x,y$ represent the coordinates of a point on the map. $(\mu_x ,\mu_y)$ and $(\sigma_x ,\sigma_y)$ are the corresponding means and standard deviations of the distribution, respectively. 

In this work, in order to generalize this phenomenon. Our model learns 16 maps following the same gaussian laws with different means ($\mu$) and standard deviations ($\sigma$). The integration of this module allows the feature extractor to only focus on modeling the stimuli-specific features without handling biases related to the saliency task.


    \subsection{Decoder}
The decoder is composed mainly of a merging module, a channel-wise selection network, and the Soft-ArgMax function. 
Our merging module network consists of 8 ($3$x$3$) convolutional layers, each activated by a ReLU function. The succession of these layers ingests the feature maps extracted by the feature extractor network concatenated with the learnable prior maps previously described. The merging network combines stimuli-specific features extracted by MobileNet and prior maps representing the learned biases models, while gradually reducing the number of feature maps to 20.

An activation probabilities vector from the channel selection network weights the resulting feature maps. More details about the selection network are given in section 2.4. The latter allows predicting fixation point coordinates from the model using the Soft-ArgMax Function (SAM) \cite{sam} in a way  similar to \cite{SALYPATH}.
The SAM function is described as follows:

\begin{equation}
  SAM(x) =  \sum_{i=0}^{W} \sum_{j=0}^{H} \frac{e^{\beta x_{i,j}}}{\sum_{i^\prime =0 }^{W} \sum_{j^\prime = 0 }^{H} e^{\beta x_{i^\prime ,j^\prime }}}({\frac{i}{W},\frac{j}{H}})^T
  \label{eq:SAM}
\end{equation}
where ${i,j,i^\prime ,j^\prime}$ iterate over pixel coordinates. ${H,W}$ represent the height and width of the feature map, respectively. $x$ is the input feature map and $\beta$ is a parameter adjusting the distribution of the softmax output.

    \subsection{Channel Selection}
    
The primary function of the channel selection layer is to create a binary mask vector that chooses the suitable fixation points to be part of the predicted scanpath. This means it selects the points suitable to be part of the scanpath and discards other predicted points using a binary mask vector. This selection mechanism allows generating variable-length scanpaths. 

The configuration takes the feature maps originating from the merging module and passes them to Global Max Pooling and Average Pooling layers in parallel after flattening. The obtained vectors are used as input to a shallow 3-layer Multi-Layer Perceptron (MLP) network ( i.e GMPN, GVPN). The resulting vectors are concatenated, then ingested by a $3^{rd}$ MLP (i.e MN) network activated by a Sigmoid in the last layer to provide a vector with values in the continuous range $[0,1]$. This vector is then binarised according to its mean value as follows:   
    


\begin{equation}
    mask_b= v > mean(v)  
    \label{eq:b_mask}
\end{equation}
with
\begin{multline}
    v = MLP_3(cat( MLP_1(GMaxPool( flat(d))) , \\ MLP_2(GAvgPool(flat(d)))))      
    \label{eq:p_mask}
\end{multline}
where $mask_b$ is the resulting binary mask vector. $MLP_1,MLP_2,MLP_3$ are the shallow MLP networks, $cat$ is a vector concatenation, $GMaxPool$  is the global max pooling operation,  $GAvgPool$  is the global average pooling operation, $flat$ is a flattening function and $d$ is the output feature maps obtained from the Merging layers.

Due to the non-differentiability of the binarization function, we cut the network's computational graph and multiply the vector $v$ by $d$ to allow the back-propagation. The Sigmoid function activates the last layer of the $MLP_3$ to avoid an exploding gradient problem and maintain operations in the same range as the network's output. 

    \subsection{Training and Loss }
    
The model was trained on 9000 images and validated on 1000 from the Salicon dataset \cite{salicon}. 
We trained the model using the following loss function : 
\begin{equation}
    L (y,\hat{y}) = BCE(y,\hat{y}) + 0.001 *  \sqrt{len(y)^2-len(\hat{y})^2} 
    \label{eq:loss}
\end{equation}
where $y$ and $\hat{y}$ are the predicted and ground truth scanpath vectors, respectively. $BCE$ is the Binary Cross-Entropy function and $len$ represents the length of the scanpath. 

The loss function is designed to take into account the difference in length between scanpaths in addition to the spatial distribution of the fixation points.  We used the Adam optimizer with a learning rate of $5x10^{-5}$ for 70 epochs and initialized all the weights using the Xavier initializer. 


    \subsection{Domain Adaptation} 


The training of our model on natural scene images yields the results shown in Sec. \ref{sec:results}. However, preliminary test results of the same model on images belonging to other source domains (i.e. paintings, etc.) did not perform as well. Therefore, to fully utilize the power of our models, we need to adapt them to a new source domain.  
The model must be able to perceive the two domains (i.e. the original natural scene and painting) as being part of the same data distribution $\mathcal{D}$ for that purpose  we use the adaptation method proposed in \cite{DA}. That is achieved by minimizing the perceived distance between the two distributions of natural images $\mathcal{D}_n$ and paintings $\mathcal{D}_p$. This means that our feature extractor should maximize the ability to identify the mutual useful information between the two domains and discard the ineffective features for the prediction task.  
    

As we train our models on images from both distributions, we add a small branch to the network which classifies the images as being from $\mathcal{D}_n$ or $\mathcal{D}_p$ in pseudo-labeled manner. 
We add a gradient reversal layer (GRL) at the start of this branch, which reverses the sign of the gradient flow during back-propagation. Eq. \ref{eq:GRL_forward} defines forward propagation while Eq. \ref{eq:GRL_backward} concerns back-propagation. Both equations are given below: 

\begin{equation}
    GRL(x) = x   
    \label{eq:GRL_forward}
\end{equation}
\begin{equation}
    GRL(\frac{\partial L}{\partial x}) = - \frac{\partial L}{\partial x}   
    \label{eq:GRL_backward}
\end{equation}
where $x$ is the input of the layer, and  $\frac{\partial L_d}{\partial x}$ represents the gradient of the domain loss $L_d$ when back propagating through the network.

While the classifier network learns to discriminate the images from the 2 domains, the reversal of the gradient sign pushes the feature extractor to find a unified representational space between the domains thus minimizing the distance between their distributions.
This forces the feature extractor to disregard the domain-specific features and noises and emphasize the mutual characteristics of the two domains.
This can be modeled as the union of the 2 distributions minus the noise distribution of each of the domains: 

\begin{equation}
     \mathcal{D} = \mathcal{D}_p + \mathcal{D}_n - ( \mathcal{N}_p + \mathcal{N}_n )  
    \label{eq:dists}
\end{equation}
where $\mathcal{D}_n , \mathcal{D}_p$ and $\mathcal{D}$ are defined as before and $\mathcal{N}_p$ and $\mathcal{N}_n$ are the specific noise distributions of the source domain. 

For unsupervised domain adaptation training, we used a mix of 2000 Salicon \cite{salicon} images and 2000 unlabeled paintings images scraped from the internet. 

\section{Experimental Results}
\label{sec:results}


\begin{table*}[h!]
\begin{center}
\begin{tabular}{ c c c c c c c c }
\hline
\textbf{Model} & \textbf{Shape } & \textbf{Direction} & \textbf{Length} & \textbf{Position} & \textbf{MM Score} & \textbf{NSS} & \textbf{Congruency}\\ 
\hline
 PathGAN \cite{pathgan} & 0.9608 & 0.5698 & 0.9530 & 0.8172 & 0.8252  & -0.2904  &  0.0825 \\ 
 \hline
 Le Meur \cite{lemeur} & 0.9505 & 0.6231 & 0.9488 & 0.8605 & 0.8457 &  0.8780 &   0.4784   \\ 
 \hline
 G-Eymol \cite{G-Eymol} & 0.9338 & 0.6271 & 0.9521 & 0.8967 &  0,8524 &  0.8727 & 0.3449 \\
 \hline
 SALYPATH \cite{SALYPATH}  & 0.9659  & \textbf{0.6275} & 0.9521 & 0.8965  &   0,8605  & 0.3472 &  0.4572     \\ 
 \hline
 Our Model  & \textbf{0.9702} & 0.6173 & \textbf{0.9587} & \textbf{0.8968} &  \textbf{0,8607} &  \textbf{1.0140} &   \textbf{0.5170} \\
 
 \hline
 
\end{tabular}

\caption{ \label{tab:MM-salicon}Results of scanpath prediction on Salicon dataset}
\vspace{-5mm}
\end{center}
\end{table*}

\begin{table*}[h!]
\begin{center}
\scalebox{0.9}{
\begin{tabular}{ c  c  c  c  c  c  c c}
\hline
\textbf{Model} & \textbf{Shape } & \textbf{Direction} & \textbf{Length} & \textbf{Position} & \textbf{MM Score} & \textbf{NSS} & \textbf{Congruency}\\ 
\hline
 PathGan \cite{pathgan} & 0.9237  & 0.5630   & 0.8929   &  0.8124    &   0.7561  & -0.2750  &  0.0209        \\ 
 \hline
 DCSM (VGG) \cite{DCSM}  & 0.8720 & 0.6420 & 0.8730 & 0.8160 & 0,8007 & - & - \\ 
 \hline
 DCSM (ResNet) \cite{DCSM} & 0.8780 & 0.5890 & 0.8580 & 0.8220 &  0,7868 & - & -\\ 
 \hline
 Le Meur \cite{lemeur} & 0.9241  &  0.6378  & \textbf{0.9171}  &  0.7749 &  0,8135  & 0.8508   &  0.1974    \\ 
 \hline
 G-Eymol \cite{G-Eymol} & 0.8885 & 0.5954 & 0.8580 & 0.7800 &  0,7805 & 0.8700 & 0.1105\\
 
 \hline
 SALYPATH \cite{SALYPATH} & 0.9363  & \textbf{0.6507} & 0.9046 & 0.7983  &   0,8225 &  0.1595  &   0.0916    \\ 
 \hline
 Our model & \textbf{0.9392}  & 0.6152 & 0.9100 & \textbf{0.8537} &  \textbf{0.82952} &  \textbf{0.8888}  &   \textbf{0.2114}   \\
\hline
\end{tabular}}
\caption{\label{tab:mit1003_results}Results of scanpath prediction on MIT1003.}
\vspace{-5mm}
\end{center}
\end{table*}

    \subsection{Datasets}
    \label{sec:datasets}
In this study, three datasets have been used to evaluate our method: Salicon \cite{salicon}, MIT1003 \cite{MIT1003}, and Le Meur paintings \cite{paint_lemeur}. 

\textbf{Salicon} is a large-scale dataset extracted from MS-COCO \cite{Coco} dataset. Each image is accompanied by the corresponding saliency map and scanpaths. Here, we used $9000$ images for the training set and $1000$ images for the validation set. We also used $5000$ images for testing and comparison with other models.
In our work, we tested approximately 250000 scanpaths on this dataset. The great number of scanpaths ensures the soundness of the empirical results.

\textbf{MIT1003} is one of the most well-known datasets for static image saliency,  and it used usually together with the MIT300 dataset \cite{MIT300}. MIT1003 is composed of 1003 natural scene images depicting objects, individuals, and scenes in multiple conditions. Each image is provided together with its saliency map and scanpaths from 15 observers, this constitutes 15045 scanpaths available for testing.   
We use the whole dataset essentially to evaluate the generalization ability of our model (i.e. cross-dataset). Meaning we did not fine-tune our model on it. 

\textbf{Le Meur paintings} \cite{paint_lemeur} is a saliency dataset for paintings, which is composed of 150 images for paintings with their saliency maps. This dataset is employed to evaluate the ability of our model to generalize the knowledge acquired from natural scene images to the paintings domain and test the effects of our domain adaptation on this dataset. 
The dataset does not provide observers scanpaths, but it is the only visual attention paintings dataset publicly available.

    \subsection{ Experimental protocol and metrics}
We used the same set of images and partitions described in section \ref{sec:datasets} to test our model and the compared methods. The results of each model were compared to the ground truth using three metrics: $MultiMatch$ \cite{Multimatch}, $NSS$ \cite{NSS} and $Congruency$ \cite{congruency}.

\textbf{Multimatch}  compares two vectors representing the scanpaths through 5 different characteristics (i.e. Shape, Direction, Length, Position and Duration).


\textbf{NSS} \cite{NSS} represents the average normalized saliency value of a fixation point from the scanpath on the saliency map, while  \textbf{Congruency} \cite{congruency}  calculates the average ratio of the number of fixations that fall   within a salient region and the global number of fixations.

    \subsection{Results}
    \label{sec:Results}
    

\begin{table}[htp!]
\begin{center}
\scalebox{1}{
\begin{tabular}{ c c c}
\hline
\textbf{Model} & \textbf{NSS} & \textbf{Congruency}  \\ 
\hline
Our model (without DA)  &  1.3620  &   0.4024   \\
\hline
Our model (with DA)  &  \textbf{1.5093}  &    \textbf{0.4244}  \\
\hline
\end{tabular}}
\caption{\label{tab:lemeur_results}Results of scanpath prediction on Le Meur paintings before and after domain adaptation.}
\vspace{-10mm}
\end{center}
\end{table}

\begin{figure*}[!ht]
    \centering
    \includegraphics[height= 60mm]{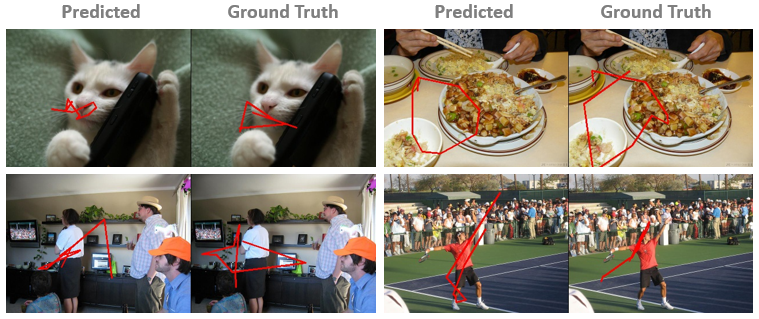}
    \caption{\label{figure:Qual_gen}Qualitative Results on Salicon Validation.}
    \vspace{-1mm}
\end{figure*}

\begin{figure*}[!ht]
   \centering
    \includegraphics[width= 0.9 \linewidth]{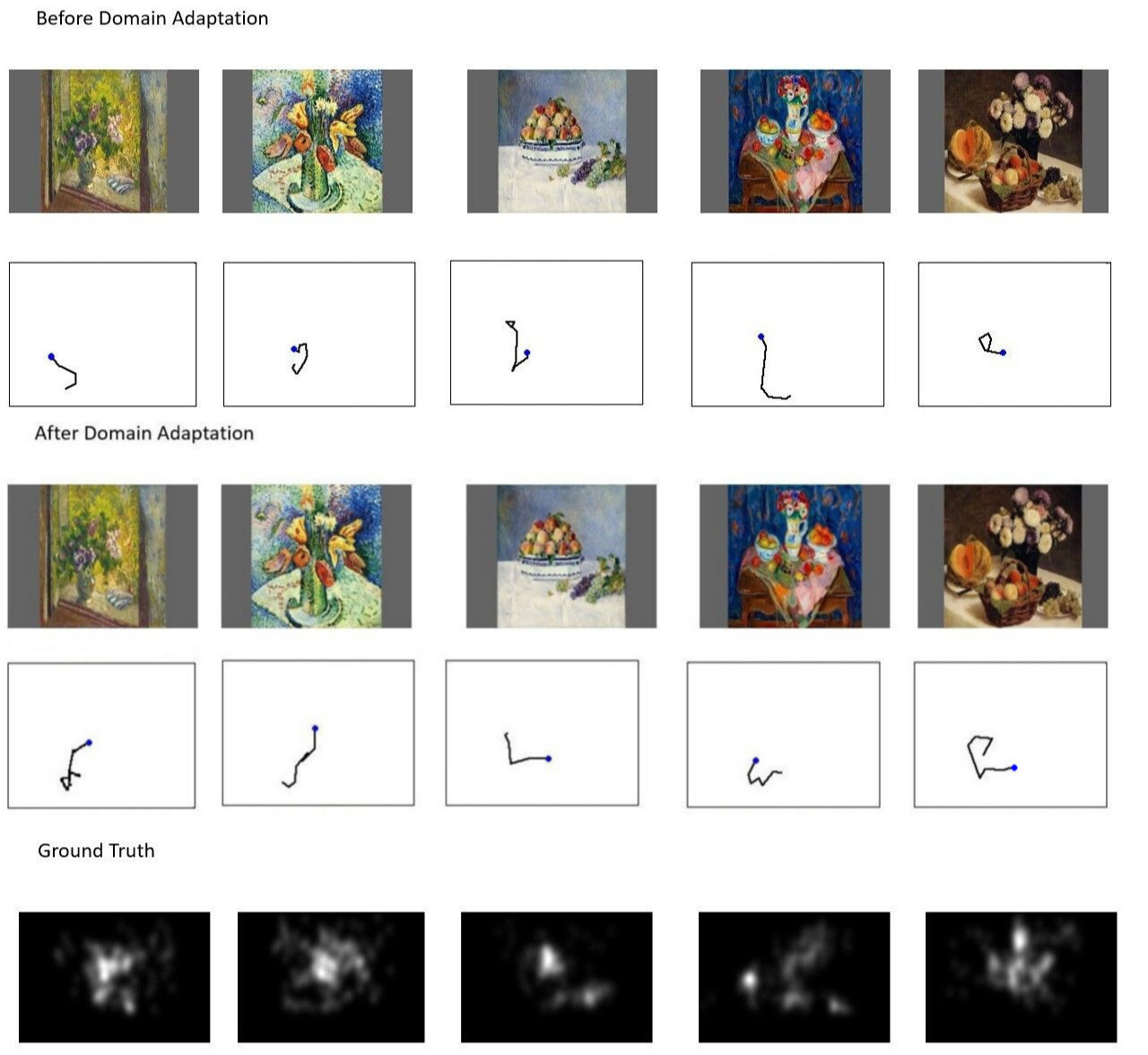}
    \caption{\label{figure:QualtDA}Predictions  on Le Meur dataset Before and After Domain Adaptation.}
    \vspace{-5mm}
\end{figure*}

The testing results on Salicon are shown in Table \ref{tab:MM-salicon}. The highest scores are highlighted in bold. As can be seen, our model achieves the best results on the $Shape$, $Length$ and $Position$ components while demonstrating acceptable results for the $Direction$ component.
An slightly surpasses the runner-up on $MM_Score$ which represents the mean score of the 4 components.  
Overall, our model's performances on $MultiMatch$ metrics are comparable to the state-of-the-art models; it even slightly surpasses SALYPATH in some criteria. As far as it concerns the $NSS$ metrics' scores, our model achieves the best results exceeding Le Meur and G-Eymol, which use saliency maps for their scanpath generation. Yet, the model is surpassed by Le Meur and SALYPATH on the $congruency$ metric. That can be explained by the fact that Le Meur heavily relies on saliency maps for scanpath generation while SALYPATH learned saliency intermediate features from the saliency prediction branch.

Table \ref{tab:mit1003_results} shows the results obtained on MIT1003. As it can be seen, our model scores the highest on the $Shape$ and $Position$ criteria while its performances on the $Direction$ and $Length$ characteristics score slightly lower albeit still competitive to those of the state-of-the-art methods. Nonetheless our model surpasses the other models mean score $MM Score$.
Our model $NSS$ and $Congruency$ scores decrease a little. Still they scored higher than the comparison models. 
In Figure \ref{figure:Qual_gen}, a qualitative depiction of the results predicted by our model is given. The figure includes the predicted scanpaths along with ground truth of scanpaths. These results show that our model can efficiently and effectively predict plausible scanpaths from images.      

    \subsection{Domain Adaptation Results}

We evaluated the performance of the domain adaptation of model using a painting dataset by Le Meur. Unfortunately, the dataset provides only stimuli images and saliency maps with no scanpaths. That prevented us from using the $MultiMatch$ metric in our evaluation. Thus we only used the hybrid metrics $NSS$ and $Congruency$. 
Table \ref{tab:lemeur_results} aims at highlighting the improvements of our model after introducing domain adaption. $NSS$ metrics' scores increased considerably and a improvement can even be noticed on the $Congruency$ results which shows the effectiveness of the domain adaptation in making a closer distance between the two different datasets. Fig. \ref{figure:QualtDA} provides a visual and qualitative demonstration of the domain adaptation to the scanpath generation on the new domain of paintings.  We can observe that the scanpaths before domain adaptation have very  small saccadic lengths and concentrate the fixations on a very small region which is sometimes not salient, this can be attributed to the effect of painting styles. On the other hand the scanpaths obtained after the  domain adaptation span wider regions and is more correlated to saliency maps depicted in the last row.       

\section{Conclusion}
\label{sec:conclusion}
In this paper, we presented a new deep learning architecture to predict scanpath for 2D images, we later adapt the approach to painting images which represent a fundamental segment of cultural heritage. Our model consists of a feature extractor reinforced task oriented learnable prior maps. The network also uses a channel-wise selection module that ensures the stochastic nature of the scanpath length between images. The proposed model was evaluated against state-of-the-art ones and produced the best results. Using an adversarial unsupervised domain adaptation algorithm, the model has later adapted from the natural scenes domain to the paintings domain, producing very relevant results. The Soft-ArgMax function creates limitation as it is not able to predict the duration of fixations, in future works we will add a time prediction module as well  as a incorporate a probabilistic generative approach in order to  make the prediction process of stochastic nature. 
Nonetheless incorporating the domain adaptation approach for scanpath prediction, opens doors for to further expand the model to other image domains related to heritage understanding like monuments architectures, statues, and artificial 3D images applications like virtual museum visits.    
\vspace{-3mm}
 \begin{acks}
Funded by the TIC-ART project, Regional fund (Region Centre-Val de Loire)
\end{acks}


\bibliographystyle{ACM-Reference-Format}
\bibliography{ref}



\end{document}